\pdfoutput=1

\documentclass[11pt]{article}

\usepackage{acl}

\usepackage{times}
\usepackage{latexsym}

\usepackage[T1]{fontenc}

\usepackage[utf8]{inputenc}

\usepackage{microtype}

\usepackage{graphicx}
\usepackage{amsmath}
\usepackage{amsmath}
\usepackage{multirow}
\usepackage{array}
\usepackage{stfloats}
\usepackage{multirow}
\usepackage{multicol}
\usepackage{graphicx}
\usepackage{bm}
\usepackage{amsfonts,amssymb} 
\usepackage{verbatim}
\usepackage{makecell}
\usepackage{setspace}
\usepackage{mathtools}

\title{To Answer or Not to Answer? Improving Machine Reading Comprehension Model with  Span-based Contrastive Learning}

\author{Yunjie Ji, Liangyu Chen, Chenxiao Dou, Baochang Ma, Xiangang Li\\
  Beike Inc., Beijing, China  \\
  \texttt{\{jiyunjie001,chenliangyu003,douchenxiao001,}  \\
  \texttt{mabaochang001,lixiangang002\}@ke.com}}

\begin{document}
\maketitle
\begin{abstract}
Machine Reading Comprehension with Unanswerable Questions is a difficult NLP task, challenged by the questions which can not be answered from passages.
It is observed that subtle literal changes often make an answerable question unanswerable, however, most MRC models fail to recognize such changes.
To address this problem, in this paper, we propose a span-based method of Contrastive Learning (spanCL) which explicitly contrast answerable questions with their answerable and unanswerable counterparts at the answer span level.
With spanCL, MRC models are forced to perceive crucial semantic changes from slight literal differences.
Experiments on SQuAD 2.0 dataset show that spanCL can improve baselines significantly, yielding 0.86\textasciitilde 2.14 absolute EM improvements. 
Additional experiments also show that spanCL is an effective  way to utilize generated questions.   
\end{abstract}

\section{Introduction}
Machine Reading Comprehension (MRC) is an important task in Natural Language Understanding (NLU), aiming to answer specific questions through scanning a given passage\cite{hermann2015teaching, cui2016attention, rajpurkar2018know}. 
As a fundamental NLU task, MRC also plays an essential role in many applications such as question answering and dialogue tasks \cite{chen2017reading, gupta2020conversational, reddy2019coqa}.
With the rapid development of pre-trained language models (PLMs), there is also a paradigm shift \cite{schick2020exploiting, dai2020effective, sun2021paradigm} reformulating other NLP tasks (e.g. information extraction) into MRC format, especially for open-domain scenarios \cite{li2019unified, yan2021unified}.

\begin{figure} [t!]
	\centering
	\includegraphics[scale=0.375]{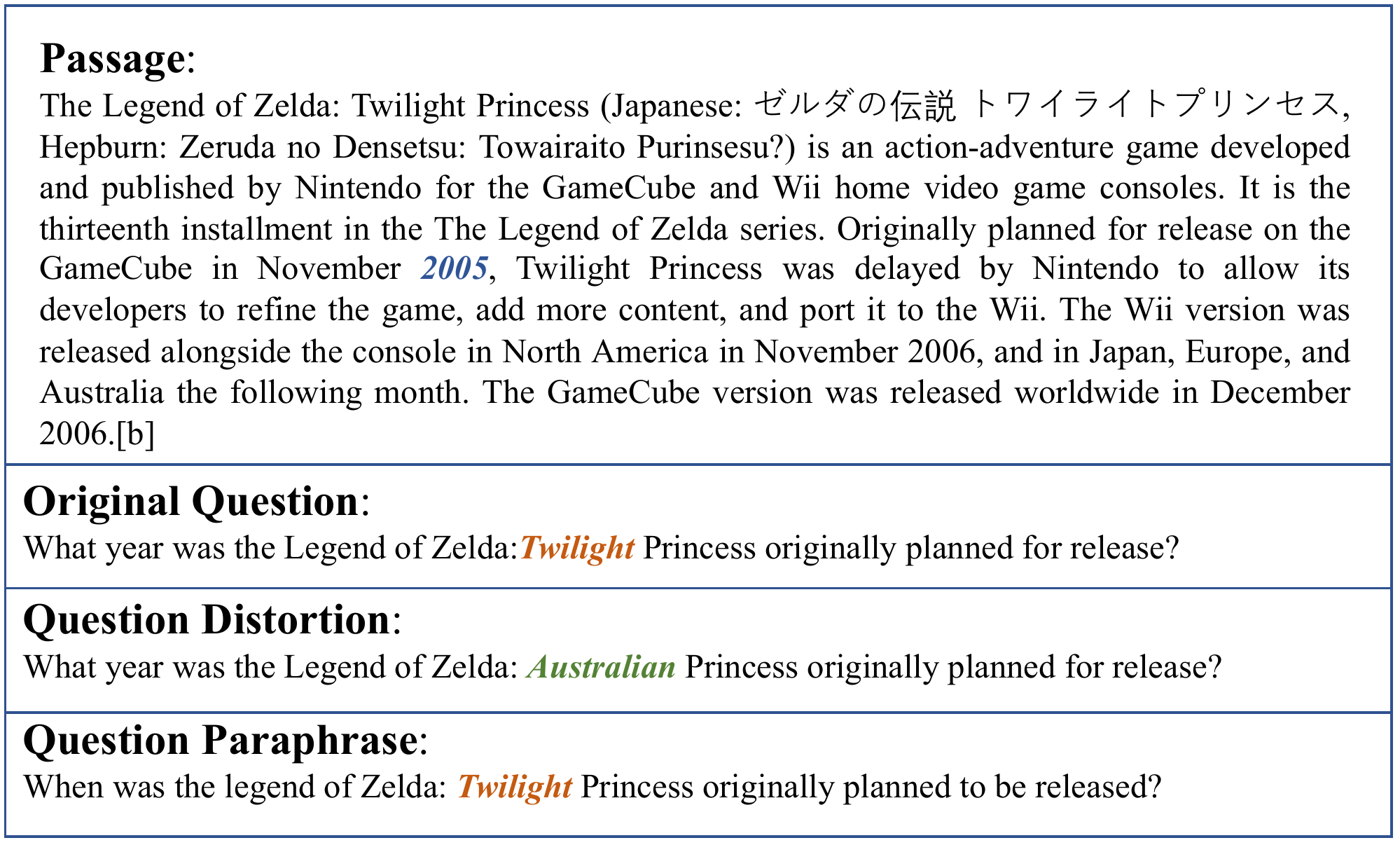}
	\caption{Question Distortion and Question Paraphrase are derived by slightly changing Original Question.}
	\label{explain}
\end{figure}

In most of the application scenarios, there exists a hypothesis that only answerable questions can be asked, which is somehow unrealistic and unreasonable.
Thus, the model that is capable of distinguishing unanswerable questions is more welcomed than the model that can only give plausible answers \cite{rajpurkar2018know}.
However, the challenge, that a slight literal change may transfer answerable questions into unanswerable ones, makes MRC models hard to gain such capability\cite{rajpurkar2018know}. 
For example, in Figure \ref{explain}, the original answerable question becomes unanswerable by only replacing \texttt{Twilight} with \texttt{Australian}, but the small literal modification towards paraphrasing does not change the answer.
Recent MRC models which predict answers using context-learning techniques and type-matching heuristics are not easy to perceive such subtle but crucial literal changes\cite{weissenborn2017making,jia2017adversarial}.
If different questions share many words in common, these models are most likely to give them the same answer, i.e., \texttt{2005} may be answered for all the three questions in Figure \ref{explain}.

To address the aforementioned challenge, we propose a span-based method of Contrastive Learning  (spanCL) in this paper.
By explicitly contrasting answerable questions with their paraphrases and their distortions, MRC models are forced to recognize the subtle but crucial literal changes.
Using pre-trained language model (PLM) as encoder, most contrastive learning methods adopt \texttt{[CLS]} as the sentence representation \cite{luo2020capt, wu2020clear, gao2021simcse, yan2021consert, wang2021cline}. 
However, in this problem, as the differences between contrastive questions are very subtle, \texttt{[CLS]} is inadequate to capture such small changes.
To solve the challenge, we propose a novel learning method, which incorporates the comparative knowledge between answerable and unanswerable questions, and exploits the semantic information of answer spans to improve the sentence representation. 
Overall, our contributions are summarized as two folds:
\begin{itemize}
\item 
To improve MRC model's capability of distinguishing unanswerable questions, we propose a simple yet effective method called spanCL, which teaches the model to recognize crucial semantic changes from slight literal differences.
\item
Comprehensive experiments show that spanCL can yield substantial performance improvements of baselines.
We also show that spanCL is an effective way to utilize generated questions.   
\end{itemize}

\section{Related Work}
\noindent \textbf{Models for MRC.} \quad 
With the help of various large-scale reading comprehension datasets \cite{hermann2015teaching, hill2015goldilocks, trischler2016newsqa, rajpurkar2016squad, lai2017race, rajpurkar2018know}, neural networks have achieved a great success on MRC in recent years. 
At first, these models are typically designed with a LSTM \cite{hochreiter1997long} or CNN \cite{lecun1998gradient} backbone, based on word embeddings \cite{mikolov2013efficient, pennington2014glove}, leveraging various attention mechanisms to build interdependent representations of passage and question \cite{ kadlec2016text, dhingra2016gated, cui2016attention, seo2016bidirectional}. 
Recently, pre-trained language models (PLMs) made a profound impact on NLP tasks \cite{radford2018improving, devlin2018bert, yang2019xlnet, liu2019roberta, lan2019albert, clark2020electra,  brown2020language, fedus2021switch}. 
With millions, billions even trillions of parameters, PLMs show a great capacity of capturing contextualized representations, and significantly boost the performance of MRC models.

\noindent \textbf{MRC with Unanswerable Questions.} \quad 
Knowing what you do not know is a crucial aspect of model intelligence \cite{rajpurkar2018know}.
In the field of MRC, a model should abstain from answering when no answer is available to the question.  
To deal with unanswerable questions, previous researchers mostly focused on designing a powerful answer verification module \cite{clark2017simple, liu2018stochastic, kundu2018nil, hu2019read+}.
Recently, a double checking strategy is proposed, in which an extra verifier is adopted to rectify the predicted answer \cite{hu2019read+, back2019neurquri, zhang2020sg, zhang2020retrospective}.
Besides the idea of designing verification modules, some other studies try to solve the problem through data augmentation, namely to synthesize more QA pairs \cite{yang2019improving, alberti2019synthetic, zhu2019learning, liu2020tell}.

\noindent \textbf{Contrastive Learning.} \quad 
To obtain rich representations of texts for down-stream NLP tasks, there have been numerous investigations of using contrastive objectives on strengthening supervised learning \cite{khosla2020supervised, gunel2020supervised} and unsupervised learning \cite{chen2020simple, gao2021simcse} in various domains \cite{he2020momentum,lin2020world,iter2020pretraining,kipf2019contrastive}.  
The main idea of contrastive learning (CL) is to learn textual representations by contrasting positive and negative examples, through concentrating the positives and alienating the negatives.
In NLP tasks, CL is usually devoted to learning rich sentence representations \cite{luo2020capt, wu2020clear}, and the main difference between these methods is the approach to find positive and negative examples. 
\newcite{wang2021cline} argued that using hard negative examples in CL is helpful to improve the semantic robustness and sensitivity of pre-trained language models.
Enlightened by the promising effects of CL, \newcite{kant2021contrast} proposed to use CL in visual question answering.
He focused on playing CL on MRC by comparing multiple answer candidates, but neglected the fact that not all questions can be answered through a given paragraph. 

\begin{figure*} [!htb]
	\centering
	\includegraphics[scale=0.5]{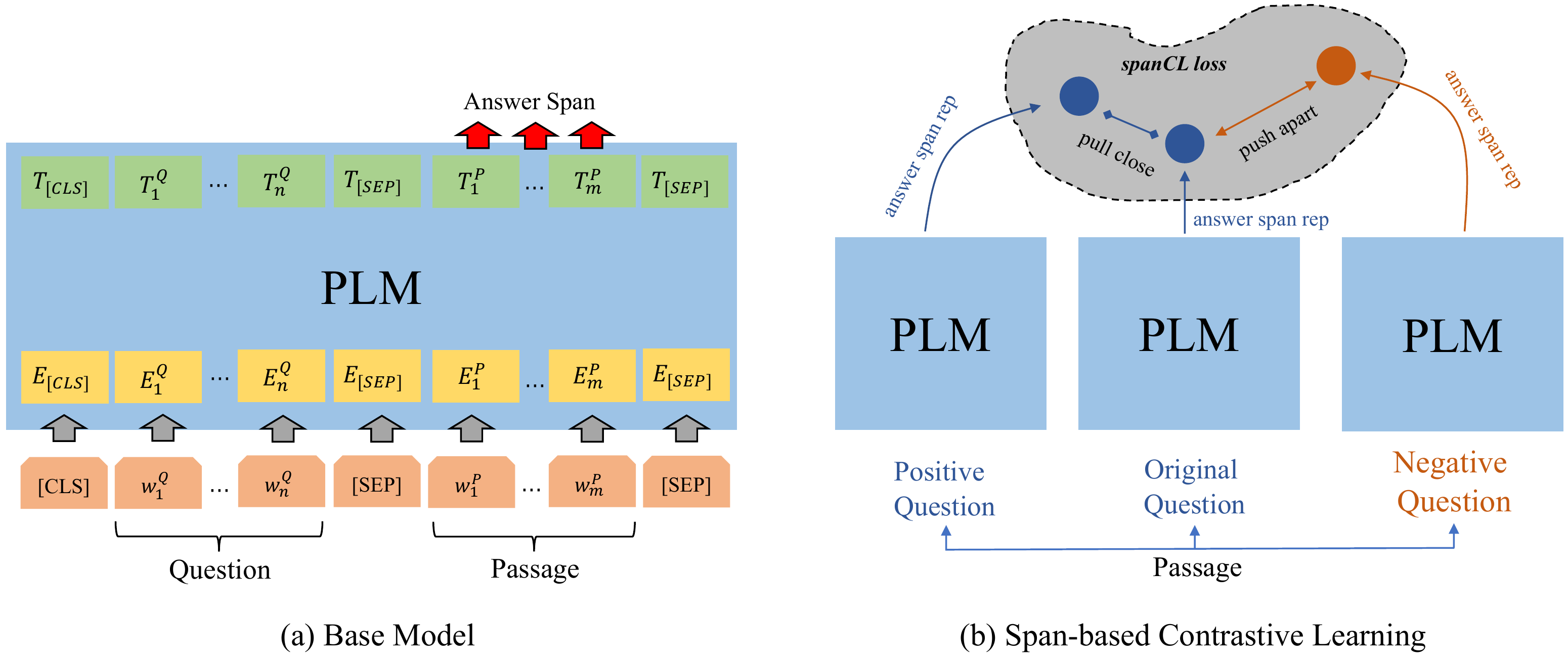}
	\caption{(a) The baseline model for MRC-U. (b) Span-based contrastive learning on answer-span representation.}
\label{model}
\end{figure*}

\section{Approach}
In this section, we first introduce the task of Machine Reading Comprehension with Unanswerable Questions (MRC-U).
Then, a baseline MRC model based on PLM is described.
At last, we propose a span-based contrastive learning method for MRC-U, named as spanCL.
In this paper, question paraphrase and positive question, question distortion and negative question are used interchangeably.

\subsection{Task Description}
In this paper, we focus on studying extractive MRC, in which the expected answer of a question is a word span of a given passage.
Thus, given a textual question $Q$ and a textual passage $P$, our goal is to find the answer span $(y_s,y_e)$ to $Q$ in $P$, where $y_s$ is the answer start position in $P$ and $y_e$ is the answer end position in $P$.

\subsection{Basic MRC Model}\label{basic_model}
We use the model same as \newcite{devlin2018bert} as the basic model for MRC-U task.
When a question and a passage are input, if the question is answerable, the model is expected to give a legal answer span $(y_s,y_e)$ in the passage; if the question is unanswerable, the model is expected to output the \texttt{[CLS]} span $(0,0)$, which indicates no related answer can be found in the passage.
The overall structure of the network is presented in Figure \ref{model}.

For illustration, we denote the output of PLM's last layer as the sequence representation, $\boldsymbol{H} \in \mathbb{R}^{l \times d}$, where $l$ is the sequence length and $d$ is the dimension.
Accordingly, the hidden representation of the $i$-th token in the sequence is denoted as $\boldsymbol{h}_i \in \boldsymbol{H}$. 
To find the start position of an answer, a start weight vector $\boldsymbol{w}_s \in \mathbb{R}^d$ is introduced to calculate the beginning possibility of each position.
Formally, the probability that the answer starts at the $i$-th token is defined as 
\begin{equation}
    p^s_i = \cfrac{\mathrm{exp}(\boldsymbol{h}_i \cdot \boldsymbol{w}_s)}{\sum\limits_{j\leq l} \mathrm{exp}( \boldsymbol{h}_j \cdot \boldsymbol{w}_s)}.
  \end{equation}
Similarly, with a end weight vector $\boldsymbol{w}_e \in \mathbb{R}^d$, the probability that the answer ends at the $i$-th token is defined as
\begin{equation}
    p^e_i = \cfrac{\mathrm{exp}(\boldsymbol{h}_i \cdot \boldsymbol{w}_e)}{\sum\limits_{j\leq l} \mathrm{exp}( \boldsymbol{h}_j \cdot \boldsymbol{w}_e)}.
\end{equation}
For learning, the cross-entropy loss on identifying the answer start and end positions is taken as the training objective as 
\begin{equation}
    \mathcal{L}_{span} = - \mathrm{log}(p^{s}_{y_s}) - \mathrm{log}(p^{e}_{y_e}),
\end{equation}
where $y_s$ and $y_e$ are the start and end positions of the true answer span. 
With the learnt model, the output answer span $(y_{s}', y_{e}')$ is predicted according to 
\begin{equation}
     (y_{s}', y_{e}') = \arg\max\nolimits_{(i,j)|i\leq j} \boldsymbol{h}_i \cdot \boldsymbol{w}_s + \boldsymbol{h}_j \cdot \boldsymbol{w}_e.
\end{equation}

\begin{table*}[t!]
\small
\begin{center}
\begin{tabular}{|l|l|} 
\hline 
\textbf{Strategy} & \textbf{Example}  \\
\hline   
Negation & \makecell[l]{Original question: What was Beyonce's role in Destiny's Child? \\ Negative question: What wasn't Beyonce's role in Destiny's Child?}  \\
\hline
Entity replacement & \makecell[l]{Original question: What native people lived in the San Diego area before the Europeans arrived? \\ Negative question: What native people lived in the San Diego area before the Mexicans \\ arrived?} \\
\hline
Antonym & \makecell[l]{Original question: What part of Gothic buildings are often found terminated with enormous  \\ pinnacles? \\ Negative question: What other part of Gothic buildings are often found terminated \\ with small pinnacles?} \\
\hline
\end{tabular}
\end{center}
\caption{Strategies used to generate negative questions.}
\label{neg_rules}
\small
\end{table*}
\subsection{Span-based Contrastive Learning}
In this section, spanCL is introduced from two aspects.
First, considering the contrastive idea of CL, we give the details about how the positive and negative examples are generated.
Second, the training objective of spanCL is presented. 

\noindent \textbf{Positive Examples.} \quad
In our method, we define the positive examples as the questions which have slight literal differences but the same answers with their original questions. 
Back Translation is an effective data augmentation method \cite{xie2019unsupervised, zhang2017mixup, zhu2019freelb}, in which a text is first translated to a target language (e.g. France) from its source language (e.g. English), and then back translated to the source language.
The final back-translated text is taken as the example of augmentation.
Thanks to Back Translation, the produced examples are lexically different but semantically same with the original example. Specifically, for each question, we first produce three question paraphrases by Back Translation using three languages.
Then we select the question that has the most literal differences with the original question as the positive question.

\noindent \textbf{Negative Examples.} \quad
In our method, we define the negative examples as the questions which have slight literal differences and not the same answers with their original questions. 
Three simple strategies are adopted to produce negative examples as the following.
\begin{itemize}
\item 
\textbf{Negation}. 
A negation word is inserted or removed from the original question. 
\item
\textbf{Antonym}. 
First, spaCy \footnote{https://github.com/explosion/spaCy}   is utilized to conduct segmentation and POS for the original question.
Then, one of the words (verbs, nouns, adjectives, or adverbs) are randomly replaced with its antonym.
\item

\textbf{Entity Replacement}. 
With an answerable question, one of its entity words is randomly placed with another entity word, which has the same entity type but does not appear in any questions.

\end{itemize}

Table \ref{neg_rules} shows several negative examples derived by these strategies.
Note that question generation is not the main topic of this paper.

\begin{spacing}{1.07}
\noindent \textbf{Span-based Contrastive Learning.}\quad
Using PLM as the encoder, \texttt{[CLS]} usually serve as the sentence representation in CL \cite{gao2021simcse,wang2021cline,yan2021consert}.
When the difference between the original question and its paraphrase or distortion is very subtle, a single \texttt{[CLS]} token is not adequate to capture the difference, making the model hard to answer such question.
Therefore, we propose to improve MRC models by contrasting these questions according to their answer-span representations. 
Specifically, given a question $Q_{org}$ and its answer span $(y_s, y_e)$, through the augmentation methods mentioned previously, we synthesize one positive question $Q_{pos}$ and one negative question $Q_{neg}$.
Based on the definition of positive examples and negative examples, $(y_s, y_e)$ is the answer span to both $Q_{pos}$ and $Q_{org}$ but not to $Q_{neg}$. 
Denote $\boldsymbol{h}^{Q_{org}}_{y_s}$ and $\boldsymbol{h}^{Q_{org}}_{y_e}$ as the representation vectors of the $y_s$-th token and $y_e$-th token in the input passage $P$ for the question $Q_{org}$, $\boldsymbol{h}^{Q_{pos}}_{y_s}$ and $\boldsymbol{h}^{Q_{pos}}_{y_e}$ as those for $Q_{pos}$, and $\boldsymbol{h}^{Q_{neg}}_{y_s}$ and $\boldsymbol{h}^{Q_{neg}}_{y_e}$ as those for $Q_{neg}$.
The concatenation of ${h}^{Q_{org}}_{y_s}$ and $\boldsymbol{h}^{Q_{org}}_{y_e}$ is used as the answer-span representation to $Q_{org}$ and denoted as $\boldsymbol{z}^{Q_{org}}$. 
Similarly, the answer-span representation to $Q_{pos}$ and $Q_{neg}$ are denoted as  $\boldsymbol{z}^{Q_{pos}}$ and $\boldsymbol{z}^{Q_{neg}}$ respectively. 
Then, our span-based contrastive loss is calculated as


\end{spacing}
\begin{small}
\begin{equation}
\begin{aligned}
    \mathcal{L}_{spanCL} = -\mathrm{log} \cfrac{\mathrm{exp}(\Phi(\boldsymbol{z}^{Q_{org}}, \boldsymbol{z}^{Q_{pos}})/\tau)}{ \splitfrac{\mathrm{exp}(\Phi(\boldsymbol{z}^{Q_{org}}, \boldsymbol{z}^{Q_{pos}})/\tau))}
    {+\mathrm{exp}(\Phi(\boldsymbol{z}^{Q_{org}}, \boldsymbol{z}^{Q_{neg}})/\tau))}}
\end{aligned}
\label{spanCL_loss}
\end{equation}
\end{small}
where $\Phi(\boldsymbol{u}, \boldsymbol{v}) = \boldsymbol{u}^{\top}\boldsymbol{v}/\left \|\boldsymbol{u}\right \|\left \|\boldsymbol{v}\right \|$ computes similarity between $\boldsymbol{u}$ and $\boldsymbol{v}$ and $\tau > 0$ is a scalar temperature parameter.
With the definition, the final objective loss of our method is presented as the following:
\begin{equation}
    \mathcal{L} = \lambda_1\mathcal{L}_{span} + \lambda_2\mathcal{L}_{spanCL}.
\end{equation}

\section{Experiments}
\subsection{Datasets and Metrics}
We evaluate our method on the well-known dataset SQuAD 2.0 \cite{rajpurkar2018know}, which covers the questions of SQuAD1.1 \cite{rajpurkar2016squad} with  new unanswerable questions written adversarially by crowdworkers to imitate the answerable ones.
Moreover, for each unanswerable question, a plausible answer span is annotated, which indicates the incorrect answer obtained by type-matching heuristics.
The training dataset contains 87k answerable and 43k unanswerable questions, and half of the examples in the development set are unanswerable.

Two official metrics are used to evaluate the model performance on SQuAD 2.0: Exact Match (EM) and  F1. 
EM is used to compute the percentage of predictions that match ground truth answers exactly.
F1 is a softer metric, which measures the average overlap between the prediction and ground truth answer at token level.
\subsection{Experimental Setup}
\noindent \textbf{MRC Model.} \quad
We adopt the model introduced in \ref{basic_model} with various PLM encoders for the MRC-U task.
Bert \cite{devlin2018bert}, RoBERTa \cite{liu2019roberta}, ALBERT \cite{lan2019albert} are selected in our experiments.
We download the  pre-trained weights from Hugging Face\footnote{https://huggingface.co/bert-base-uncased}.

\noindent \textbf{Training Data Construction.} \quad
For each original answerable question, we use Back Translation to generate its paraphrase.
In SQuAD 2.0, we can find the negative questions for 18,541 answerable questions in the original dataset.
For the rest 68,280 answerable questions, we use our augmentation strategies to generate negative questions.

During our training, the span loss is calculated based on $Q_{org}$ and $Q_{neg}$.
In section \ref{main_result}, we will explain why $Q_{pos}$ is discarded for calculating span loss.


\noindent \textbf{Hyper-parameters.} \quad
We use the default hyper-parameter settings for the SQuAD 2.0 task.
Specifically, we set maximum sequence length, doc stride, maximum query length and maximum answer length to 512, 128, 64 and 30.
For fine-tuning our model, we set the learning rate, batch size, training epoch and warm-up rate to 2e-5, 12, 2 and 0.1.
The temperature in spanCL is set to 0.05.
The weights of span loss and spanCL loss are $\lambda_1=\lambda_2=0.5$. 
For each time, we fix the random seed, ensuring our results are reproducible.
We run our experiments on two Tesla A100 40G GPUs with 5 GPU hours to train a model.

\subsection{Main Results}\label{main_result}

\begin{table}[t!]
\small
\begin{center}
\begin{tabular} {l|cc|cc}
\hline 
\multirow{2}{*}{\textbf{Model}} & \multicolumn{2}{c|}{\textbf{Dev}} & \multicolumn{2}{c}{\textbf{$\Delta$}}  \\
\cline{2-5}
& \textbf{EM} & \textbf{F1} & \textbf{EM} & \textbf{F1} \\
\hline
$BERT_{base}$ & 73.37 & 76.34&-&- \\
\quad + spanCL &  75.51 & 78.34 & +2.14 & +2.00 \\
\hline
$BERT_{large}$ & 78.88 & 81.85 & - & - \\
\quad + spanCL &  79.76 & 82.61 & +0.88 & +0.76 \\
\hline
$RoBERTa_{base}$ & 78.85 & 81.42 & - & - \\
\quad + spanCL &  80.18 & 82.84 & +1.33 & +1.42 \\
\hline
$RoBERTa_{large}$ & 86.12 & 88.88 & - & - \\
\quad + spanCL &  86.98& 89.70 & +0.86 & +0.82 \\
\hline
$ALBERT_{base}$ & 77.84 & 81.27 & - & - \\
\quad + spanCL & 79.52 & 82.97 & +1.68 & +1.7 \\
\hline
$ALBERT_{large}$ &79.99 & 83.27 & - & - \\
\quad + spanCL & 81.51 & 84.67 & +1.52 & +1.4 \\
\hline
\end{tabular}
\end{center}
\caption{Results (\%) on the dev set  of SQuAD 2.0.}
\label{main_table}
\end{table}

\begin{table}[t!]
\small
\begin{center}
\begin{tabular} {l|cc|cc}
\hline 
\multirow{2}{*}{\textbf{Model}} & \multicolumn{2}{c|}{\textbf{EM}} & \multicolumn{2}{c}{\textbf{Dev}}  \\
\cline{2-5}
& \textbf{HasAns} & \textbf{NoAns} & \textbf{EM} & \textbf{F1} \\
\hline
$BERT_{base}$ & 70.31 &74.76& 73.37& 76.34 \\
\quad + pos &  \textbf{72.57} & 68.93 & 72.26 & 75.22 \\
\quad + neg &  67.05 &  \textbf{87.74} & 74.02 & 76.38 \\
\quad + pos\&neg &  66.16 & 78.48 & 72.59 & 75.37 \\
\hline
\quad + spanCL & 72.52 & 75.91 &  \textbf{75.51} &  \textbf{78.34} \\
\hline
\end{tabular}
\end{center}
\caption{Training with spanCL vs Training with expanded datasets.}
\label{training_datasets_table}
\end{table}
From Table \ref{main_table}, we notice that spanCL improves the performance of each baseline model, yielding 0.86\textasciitilde 2.14 absolute EM improvement and 0.76\textasciitilde2.0 absolute F1 improvement, demonstrating spanCL is model-agnostic and effective. 

As additional training data (i.e. the extra positive and negative questions) is used, it is necessary to analyze if the improvements are merely brought by this additional data.
We conduct experiments by training with different datasets and display the results in Table \ref{training_datasets_table}.
$\mathrm{BERT}_{base}$ means training $\mathrm{BERT}_{base}$ with original SQuAD 2.0 training set.
``$\mathrm{+pos}$'' and ``$\mathrm{+neg}$''  mean expanding the original training set with generated positive questions and generated negative questions respectively.
Surprisingly, Simply expanding the training set can not guarantee the performance improvement.
We find that  adding positive examples into the training set does not improve the performance of MRC model.  
One possible reason is that the positive questions make the model over insensitive and ignore  slight literal changes, which is inappropriate for MRC-U task. 
By comparing $\mathrm{BERT}_{base}$ with ``$\mathrm{+neg}$'', we find that training with more negative examples, the model tends to predict more \texttt{NoAns} and achieve a high performance on \texttt{NoAns}, while the performance on the \texttt{HasAns} drops a lot and the overall improvement of EM is much less than ``$\mathrm{+spanCL}$''.
From the results in Table \ref{training_datasets_table}, we can conclude that spanCL is effective to utilize the generated questions.

\subsection{Influence of Negative Examples}\label{quality_neg}
The unanswerable questions generated by our strategies are rather plain.
We believe spanCL can further boost the performance by high-quality unanswerable questions.
\newcite{liu2020tell} proposed a context-relevant generation method called CRQDA, which generates delicate negative questions \footnote{https://github.com/dayihengliu/CRQDA}. 
In table \ref{diff_negs_table}, ``+CRQDA'' denotes training the baseline model with the dataset including the  delicate negative questions generated by CRQDA.
``+spanCL with simple negatives`` denotes applying spanCL with negative questions generated by our three strategies.
``+spanCL with CRQDA'' denotes applying spanCL with negative questions generated by CRQDA.
Comparing ``+spanCL with simple negatives`` with ``+spanCL with CRQDA``,  we find that spanCL can further boost the performance by delicate negative questions.

\begin{table}[t!]
\small
\begin{center}
\begin{tabular} {l|c|c}
\hline 
\textbf{Model} & \textbf{EM} & \textbf{F1} \\
\hline
$BERT_{base}$  & 73.37& 76.34 \\
\quad + CRQDA \cite{liu2020tell} & 75.80 & 78.70 \\
\quad + spanCL with simple negatives  & 75.51 & 78.34 \\
\quad + spanCL with CRQDA & \textbf{76.12} &  \textbf{79.09}\\
\hline
\end{tabular}
\end{center}
\caption{Performance of spanCL with different synthetic negative questions.}
\label{diff_negs_table}
\end{table}

\subsection{Influence of Temperature}
\begin{figure} [t!]
	\centering
	\includegraphics[scale=0.4]{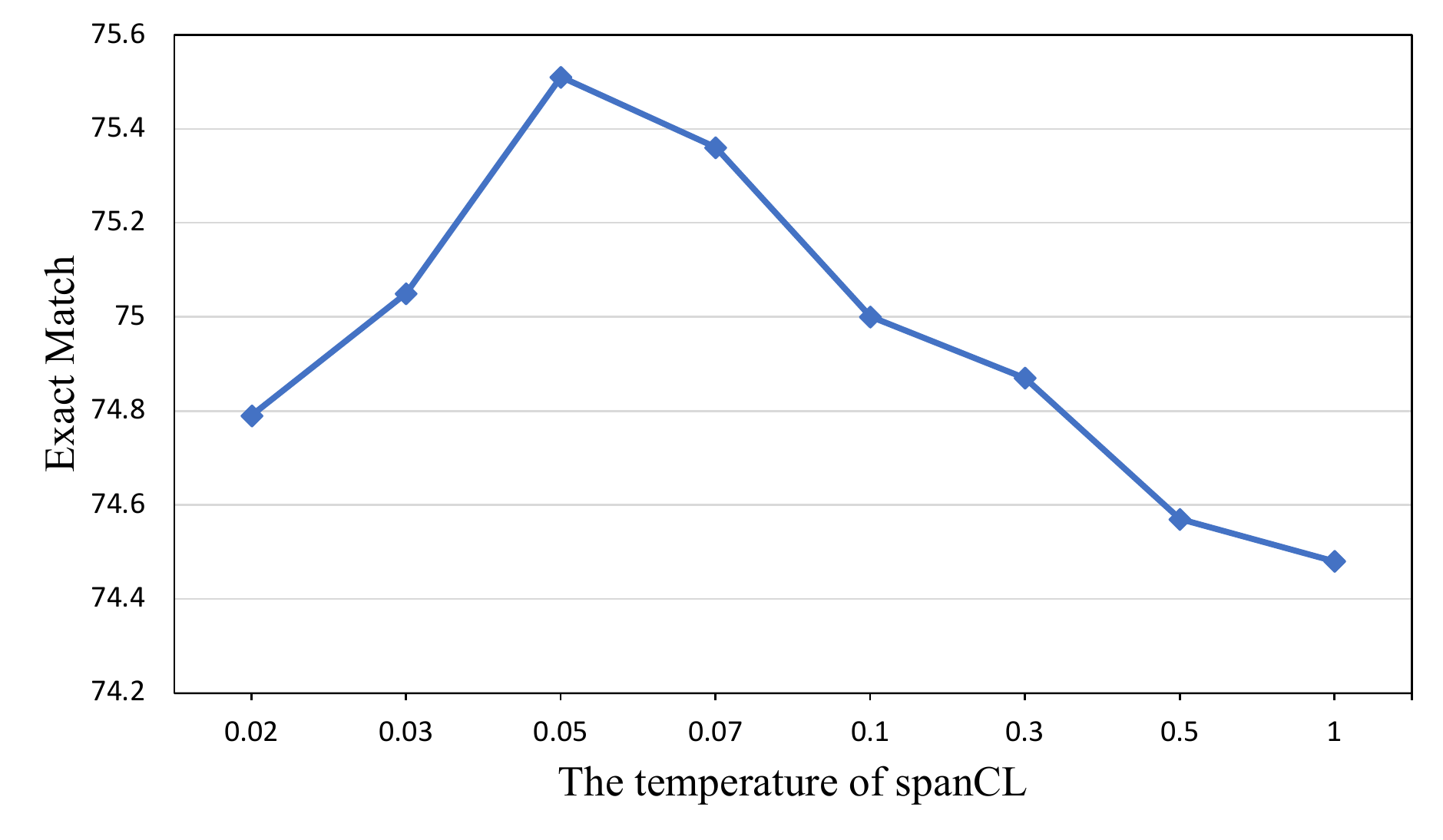}
	\caption{The influence of different temperatures in spanCL. The best performance is achieved when the temperature is set to 0.05. $\mathrm{BERT}_{base}$ is adopted as the base model.}
\label{temp_fig}
\end{figure}
The temperature $\tau$ in spanCL loss (Equation \ref{spanCL_loss}) is used to control the smoothness of the distribution normalized by the softmax operation. 
A large temperature smoothes the distribution while a small temperature sharpens the distribution.
As shown in the Figure \ref{temp_fig}, spanCL is  sensitive to the temperature value. 
In general, small temperature results in better performance. 
A practical temperature can be obtained within a small range (from about 0.02 to 0.1). 
We select 0.05 as the temperature in our experiments.

\begin{figure*} [t!]
	\centering
	\includegraphics[scale=0.57]{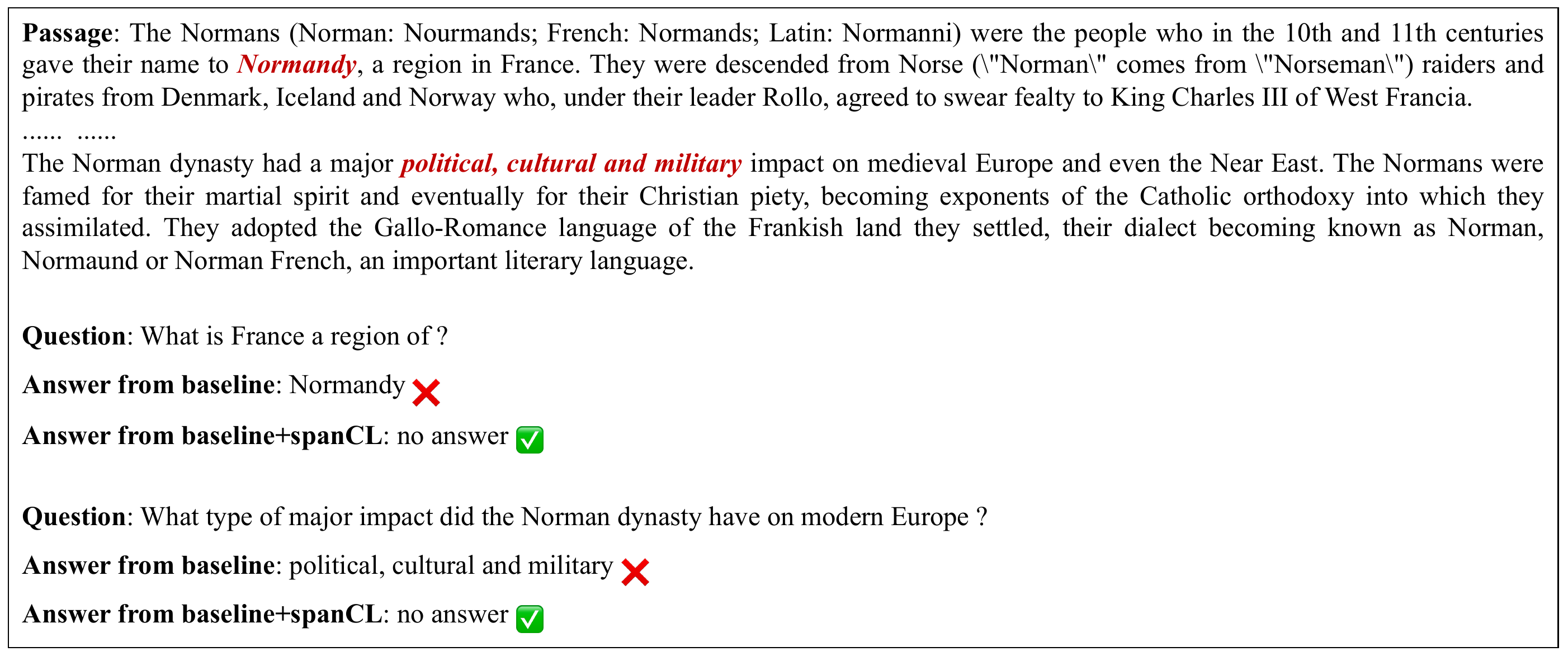}
	\caption{Qualitative Examples.}
	\label{case_study}
\end{figure*}

\subsection{Selection of Question Representations}
\begin{table}[t!]
\small
\begin{center}
\begin{tabular} {l|c|c}
\hline 
\textbf{Model} & \textbf{EM} & \textbf{F1} \\
\hline
$BERT_{base}$  & 73.37& 76.34 \\
\quad + spanCL   & \textbf{75.51} & \textbf{78.34} \\
\quad + CL with [CLS] reps & 74.18 & 77.05 \\
\quad + CL with span and [CLS] reps & 73.82 & 76.86 \\
\hline
\end{tabular}
\end{center}
\caption{Results (\%) with  different question representations used in the contrastive learning.}
\label{diff_reps_table}
\end{table}
\begin{table}[t!]
\small
\begin{center}
\begin{tabular}{l|c|c|c} 
\hline 
\textbf{Base Model} & \textbf{Training schemes} & \textbf{EM}  & \textbf{F1} \\
\hline   
$BERT_{base}$ & Joint  & \textbf{75.51}   &  \textbf{78.34}  \\
$BERT_{base}$ & Alternate & 74.67  & 77.08\\
$BERT_{base}$ & pre-train then finetune & 72.19  & 74.87\\
\hline
$BERT_{large}$ & Joint  & \textbf{79.76}     &  \textbf{82.61}  \\
$BERT_{large}$ & Alternate & 79.50  & 82.55 \\
$BERT_{large}$ & pre-train then finetune & 77.77  & 80.30\\
\hline
\end{tabular}
\end{center}
\caption{Results (\%) with  different training schemes.}
\label{training_scheme}
\small
\end{table}
In this paper, we argue that the answer span representation is better than \texttt{[CLS]}.
We conduct experiments with different question representations in this section.
When applying CL with \texttt{[CLS]} representations, we add a classification layer on the top of \texttt{[CLS]} to determine if a question is answerable or not \cite{zhang2020retrospective}, making the representation of \texttt{[CLS]} acquire the information of the question's answerability.
We also play CL with both \texttt{[CLS]} and answer-span representations, in which two CL losses are calculated together.
From Table \ref{diff_reps_table}, we can see that \texttt{CL with [CLS] reps} improves the model performance but the improvement is small than that from spanCL, and the combination of the two CL losses can confuse the model and result in a little improvement.

\subsection{Comparison between Different Training Schemes}
There are three training schemes to combine the span loss and spanCL loss: 1) joint training, in which these two losses are used together in each training step; 2) alternate training, in which the model is updated with spanCL loss after every $M$ updates with span loss; 3) pre-train and fine-tune, in which we first pre-train the model with spanCL loss and then fine-tune it with span loss.
For alternate training, we select $M$ from $\{1, 2, 3\}$ and find $M=2$ gives the best results.
From Table \ref{training_scheme}, we conclude that joint training gives the best performance and alternate training performs a little worse.
Surprisingly, with the pre-train and fine-tune training scheme, the model performs worse than the baseline model.
We guess this is because without the supervision of answer-span knowledge, it is hard to learn useful question representations.

\subsection{Qualitative Analysis}
We qualitatively analyze two representative unanswerable questions in Figure \ref{case_study}.
It can be seen that the baseline model predicts a plausible answer for each question while the baseline model trained with spanCL abstain from answering.

To correctly answer the first question, the model is asked to learn the question's semantics in sentence level.
To correctly answer the second question, the model is asked to recognize the literal change in word level. 
SpanCL can help the model perceive such crucial differences between the question and passage from both semantic and lexical aspects, and thus enable the baseline model to abstain from answering for these two questions.

\section{Conclusion}
In this paper, we propose a span-based method of Contrastive Learning (spanCL) to solve the MRC task with Unanswerable Questions.
SpanCL is devised based on the fact that an answerable question can become unanswerable with slight literal changes.
By explicitly contrasting an answerable question with its paraphrase and distortion at the answer span level, MRC models can be taught to perceive subtle but crucial literal changes.
Experimental results demonstrate that spanCL is model-agnostic and can improve MRC models significantly. 
Additional experiments show that spanCL is more effective to utilize the generated questions than other methods.
In addition, it should be noticed that how to generate high-quality question examples is not fully investigated in this paper, which may introduce a performance bottleneck to spanCL. 
Therefore, a study on question generation compatible with spanCL is  encouraged in the future.

\bibliography{custom}
\bibliographystyle{acl_natbib}




\end{document}